\documentclass[letterpaper, 10 pt, conference]{ieeeconf}  

\IEEEoverridecommandlockouts                              

\usepackage{cite}
\usepackage{amsmath,amssymb,amsfonts}
\usepackage{algorithmic}
\usepackage{graphicx}
\usepackage{textcomp}
\usepackage{xcolor}
\def\BibTeX{{\rm B\kern-.05em{\sc i\kern-.025em b}\kern-.08em
    T\kern-.1667em\lower.7ex\hbox{E}\kern-.125emX}}

\usepackage{graphicx}
\usepackage{amsmath}
\usepackage{amssymb}
\usepackage{graphicx}   
\usepackage{booktabs}  
\usepackage{hyperref}

\graphicspath{{./figs/}}
\DeclareGraphicsExtensions{.pdf, .png, .jpg,}
\usepackage{multirow}
\usepackage[utf8]{inputenc}
\usepackage{subfig}
\usepackage{graphicx}
\usepackage{balance}

\hyphenchar\font=-1 

\usepackage{color}
\usepackage[normalem]{ulem} 
\definecolor{visnecurugu}{rgb}{0.6,0.1,0.6}

\title{\LARGE \bf FINO-Net: A Deep Multimodal Sensor Fusion Framework \\ for Manipulation Failure Detection 
}

\author{Arda Inceoglu$^{1}$, Eren Erdal Aksoy$^{2}$, Abdullah Cihan Ak$^{1}$, Sanem Sariel$^{1}$ 
\thanks{$^{1}$Artificial Intelligence and Robotics Laboratory, Faculty of Computer and Informatics Engineering,
      Istanbul Technical University, Maslak, Turkey 
      {\tt\small {\{inceoglua, akab, sariel\}@itu.edu.tr}}%
      }
\thanks{$^{2}$School of Information Technology, Center for Applied Intelligent Systems Research, Halmstad  University, Halmstad, Sweden
    }
\thanks{  Arda Inceoglu was supported by the Turkcell-Istanbul Technical Uni. Researcher Funding Program. We   acknowledge the support of NVIDIA Corp. with the donation of the Quadro P6000 GPU used for this research. This research is also supported by a grant from the Scientific and Technological Research Council of Turkey (TUBITAK), Grant No. 119E-436.}
}

\begin{document}
\maketitle
\bstctlcite{IEEEexample:BSTcontrol}

\begin{abstract}
We need robots more aware of the unintended outcomes of their actions for ensuring safety. This can be achieved by an onboard failure detection system  to monitor and detect such cases. Onboard failure detection is challenging with a limited set of onboard sensor setup due to the limitations of sensing capabilities of each sensor. To alleviate these challenges, we propose FINO-Net, a novel multimodal sensor fusion based deep neural network to detect and identify manipulation failures. We also introduce FAILURE, a multimodal dataset, containing 229 real-world manipulation data recorded with a Baxter robot. Our network combines RGB, depth and audio readings to effectively detect failures. Results indicate that fusing RGB with depth and audio modalities significantly improves the performance. FINO-Net achieves \%98.60 detection accuracy on our novel dataset. Code and data are publicly available at \href{https://github.com/ardai/fino-net}{https://github.com/ardai/fino-net}.

\end{abstract}



\section{Introduction}

The use of service robots in domestic environments raises some important ethical concerns which must be taken into account in robot designs. One crucial concern is execution safety in such environments where robots work with humans and everyday objects. Although robots are equipped with well designed actions, unintended or unsafe situations may arise from failed executions of these actions in the real world due to either faults in perception, failures in actuation, or unforeseen events. Physical interactions with objects may result in collisions, fall downs, splits, or overturnings. 
Robots should monitor these types of interactions to protect objects in interaction, humans in collaboration, themselves, and  their  environments.

\begin{figure}[!t]
\centering
    \includegraphics[width=0.8\linewidth]{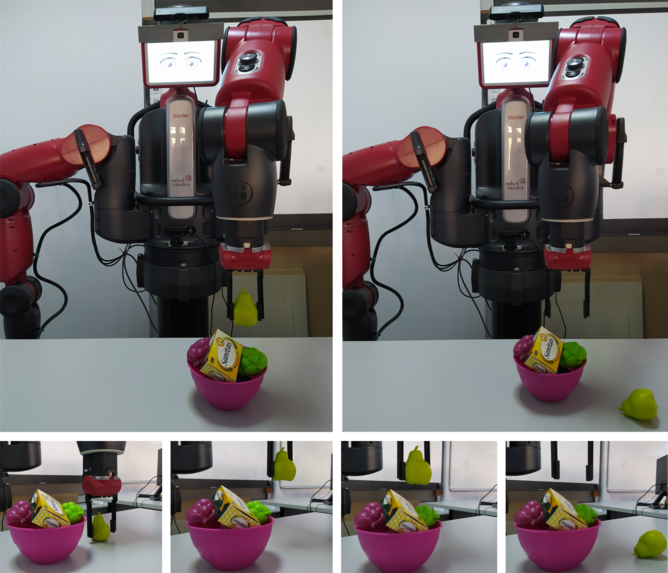}
    \caption{Snapshots from a failed place-in-container execution.}
    \label{fig:sample_scenario}
\vspace{-4mm}
\end{figure}

The primary step in execution safety is to be equipped with essential skills to detect failures or  unsafe situations. After then, predictions and precautions are possible. Execution monitoring and safety have been    well investigated research areas \cite{Fritz2005, Pettersson2005} and several methods exist for both ground and aerial robots \cite{Goel2000, Fourlas2014, Stavrou2015}. However, there still exist several research questions for service robots working in unstructured environments. This is the main motivation behind our work. We investigate how robots can effectively detect manipulation failures as a first step in safe execution. We define a \textit{failure} as an unintended outcome of an action execution. Sample failures that we focus on are tabletop manipulation actions: \textit{push}, \textit{pick\&place}, \textit{place-in}, \textit{put-on},  and \textit{pour}. A sample \textit{place-in} execution failure of our humanoid robot is given in   Fig.~\ref{fig:sample_scenario}. In this sample, the robot fails in putting a plastic pear into a bowl full of other items.     

It is almost impossible to detect and identify failures without perception. Furthermore, multi sensory integration is more desirable to take advantages of each sensor's perceptual contribution \cite{Ersen2017}. Multimodal sensor fusion methods are extensively used for manipulation, and execution monitoring also benefits from the use of multimodal data \cite{Kapotoglu2014, Park2018, Park2019, Saltali2016}. However, there exist potential challenges to deal with multimodal sensor fusion: (i) different data formats, (ii) different operating regimes/frequencies, (iii) inter and/or intra conflicting observations of sensors.

To address the above mentioned challenges, we propose a deep multimodal sensor fusion network for manipulation failure detection, named 
Failure Is Not an Option (FINO)-Net. 
FINO-Net effectively detects  manipulation failures by fusing visual (RGB and depth) and audio modalities.
Our network also adopts early fusion to combine RGB and depth frames while employing late fusion to combine vision and audio data. 
To capture spatio-temporal features in sensory observations, modalities are processed individually with a series of convolutional and convolutional-LSTM layers, then the latent space representations are  combined for detecting possible failures.

Our earlier work involves an analysis on contributions of different sensor modalities on failure detection \cite{Inceoglu2018icra}. Based on this analysis, we developed a
  Hidden  Markov  Model-based  failure detection system where the features are collected through classifiers \cite{Inceoglu2018iros}. 
 Although our new FINO-Net model uses the same modalities as input, it does not rely on precomputed features. Instead, FINO-Net  processes raw sensory data in an end-to-end manner for failure detection. 
  
The main contributions of this paper are as follows:
\begin{itemize}
    \item We introduce a novel deep neural network model that fuses multimodal sensory readings to detect   possible failure types. 
    \item We further analyze the unique contribution of each sensing modality on the failure detection task. 
    \item We release a multimodal (RGB, depth and audio) real-world dataset, named FAILURE, with $5$ different manipulation types and in total 229 manipulation scenarios.
    We evaluate the performance of FINO-Net on this dataset and compare with  a VGG-based model.

\end{itemize}


\section{Related Work}
\label{sec:related_work}

We first review the earlier literature on failure detection, execution monitoring, and multimodal sensor fusion. We then discuss some previously published manipulation datasets and elaborate on why we need to introduce our own new dataset.

\subsection{Failure Detection and Execution Monitoring}

In the literature, \textit{failure (a.k.a fault or anomaly) detection} and \textit{execution monitoring} keywords are used interchangeably. The work in \cite{Chalapathy2019} summarizes deep learning based anomaly detection for various application domains. From the robotics perspective, there are model-based and model-free approaches proposed for execution monitoring \cite{Gertler1998}. The former approaches  compare the already known models with observations, whereas the latter use sensory observation to make predictions \cite{Fritz2005, Pettersson2005}.

Among earlier model-based approaches, Kalman Filter (KF) \cite{Goel2000}, kinematic model \cite{Fourlas2014}, and residual \cite{Stavrou2015} based systems are proposed to detect and identify mechanical and sensor failures. Execution models \cite{Mendoza2014} and stochastic action models \cite{Mendoza2015} are used to detect and  correct \cite{Mendoza2019} execution anomalies. In \cite{Micalizio2013},  action models are extended to detect and recover from failures by repairing the plan.
Additionally, Description Logic (DL) \cite{Bouguerra2007}, Temporal Action Logic (TAL) \cite{Doherty2009} and Metric Temporal Logic (MTL) \cite{Kapotoglu2014} formulas are defined for execution monitoring. Furthermore, the work in \cite{Adam2014} uses a domain-specific language based approach to monitor software components. 

On the other hand, model-free execution monitoring methods employ pattern recognition techniques \cite{Pettersson2007}.
The recent work in \cite{Mauro2019} extends planning with a vision-based execution monitoring system to search for target objects in the scene to ensure planning pre-conditions are satisfied. In \cite{Sathish2019},  different preprocessing techniques for introspective data are analysed to detect gearbox failures for industrial robots. Non-parametric Bayesian models are also used for execution monitoring in robotics \cite{Zhou2020}. 
Non-parametric Hidden Markov Models (HMMs) representing spatio-temporal dynamics of anomalies are applied in \cite{Wu2019} to detect and classify anomalies by considering only introspective data (i.e., force-torque, velocity, and tactile).

The method is evaluated on pick\&place tasks by using 6 varying sized and shaped objects on a Baxter robot. A multimodal execution monitoring system is proposed for the assistive feeding task in \cite{Park2018}. The authors adopt LSTM-based variational autoencoders to process multimodal input from a sensor set including a camera, a microphone, a joint encoder and a force sensor. 

In their earlier work, multivariate Gaussian HMMs are     explored \cite{Park2019}.

Our work differs from the existing studies as we address the problem of detecting 
manipulation failures which are mainly caused by uncertainties in perception and execution. Unlike the works in \cite{Park2018,Park2019}, where the focus is rather on the failures emerging during human-robot interaction (e.g., failures due to collisions, face occlusions, utensil misses or sound from the user, etc.), we investigate the robot-object interaction failures observed over the course of an object manipulation.
Instead of hand-crafted features such as gripper status, audio events, and object displacements used in \cite{Inceoglu2018icra,Inceoglu2018iros} or sound energy, spoon position, and mouth position used in \cite{Park2018,Park2019}, our proposed perception framework learns feature representations directly from the raw multimodal sensory data in an end-to-end fashion.

\subsection{Multimodal Sensor Fusion}

Multimodal fusion has a wide variety of applications with different kinds of input modalities \cite{Ramachandram2017multimodal}. The work in \cite{Eitel2015multimodal} fuses RGB and depth images for object recognition.  RGB-D and audio data are combined in \cite{Pieropan2014audio} for human action recognition. In \cite{Kapotoglu2014}, RGB-D camera, sonar, microphone and tactile sensor data are integrated, via high level predicates extracted from sensory data, in order to detect different kind of execution failures for mobile robots.

 One of the main challenges in multimodal sensor fusion is determining when to combine modalities. Prior works introduce early, intermediate, and late fusion based architectures. Early fusion combines low-level features (i.e., raw observations), whereas late fusion incorporates high-level features (i.e., close to prediction). Intermediate architectures, where  early and late fusions are intertwined,  cover different strategies to fuse intermediate-level features. Variety of architectures are summarized in \cite{Feng2020multimodal}. In this work, we adopt a combination of early and late fusion methods to combine different sensing modalities.

\subsection{Manipulation Datasets}
Recently, there have been great efforts to collect large scale robot manipulation data. 
These efforts, however, center around manipulation skill learning tasks and ignore failures emerging during manipulation executions. For example, RoboTurk \cite{Mandlekar2018roboturk, Mandlekar2019roboturk} is a crowdsourced simulation tool to collect teleoperated manipulation data for imitation learning. The dataset also includes real robot RGB-D frames for the following tasks: object search, tower stacking and laundry layout. RoboNet \cite{Dasari2019robonet} provides 15 million video frames, from 7 different robot platforms for push and pick\&place tasks. 
In addition, the work in \cite{Levine2018learning} presents a large scale grasping dataset with a multi robot platform, and  in \cite{Finn2016unsupervised} a pushing dataset is introduced. To the best of our knowledge, there exists no publicly available multimodal datasets on robot manipulation failures. The  FAILURE dataset introduced here, as the first of its kind, rather focuses on the failed attempts of various robot manipulation actions (e.g., push, pour, pick\&place, etc.) while collecting multimodal sensor readings such as RGB images, depth data, and audio waves.


\begin{figure}[!b]
\centering
    \includegraphics[width=0.6\linewidth]{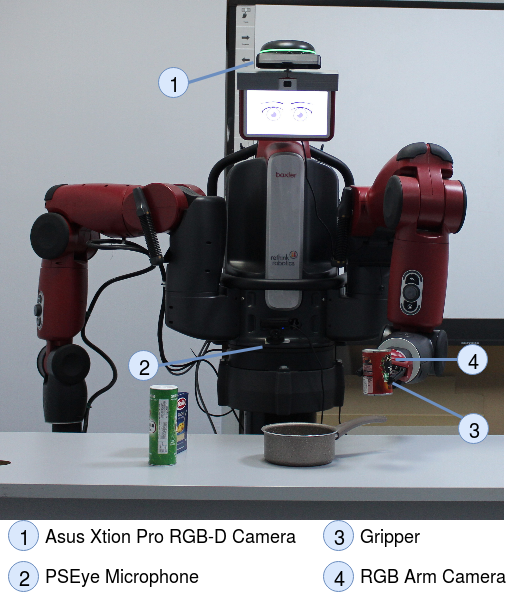}
    \caption{Experimental environment}
    \label{fig:baxter}
    \vspace{-4mm}
\end{figure}

\section{Method}

In this section, we first  describe the problem, then present the details of FAILURE dataset and FINO-Net architecture.

\subsection{Problem Description}
\label{sec:problem_desc}

In order to detect
manipulation failures, the changes in the scene should be monitored using multiple sensory modalities \cite{Inceoglu2018iros} \cite{Inceoglu2018icra}. In this work, we model the failure detection task as a classification problem as follows. 
Let there be $M \in \{1,2,...\}$ sensing modalities where $m \in M$ is modality index. Let also $D$ be the dataset containing $|D| = N$ multimodal observation sequences, $x_{t_m,i}^m$ is observation, $t_m$ is the time index for modality $m$, $i$ is the recording index and $y \in \{success, fail\}$ is the class label:
\begin{equation}
D = \{ \{(x_{1,i}^{(m)},...,x_{t_m,i}^{(m)})\}_{m=1}^M, y_i\}_{i=1}^{N}
\end{equation}

The goal is to detect failures by learning a function $\Phi(\cdot)$ that maps multimodal sensory data  to a label as either success or failure.

\subsection{Setup and Data Collection}
\begin{figure}[!t]
\centering
    \includegraphics[width=0.6\linewidth]{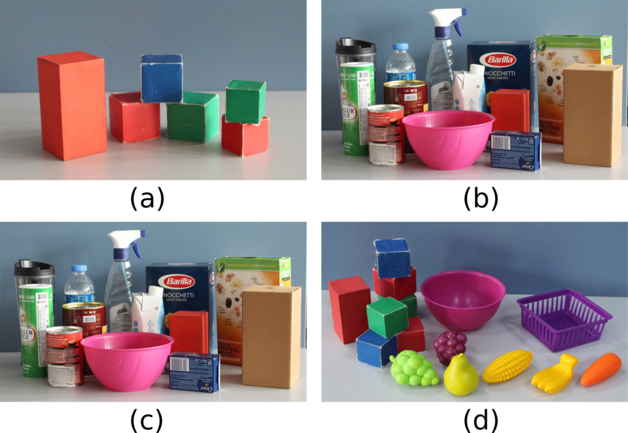}
    \caption{Object sets for the (a) Pick\&Place and Put-on-top, (b) Push, (c) Place-in-container, (d) Pour manipulations.}
    \label{fig:objects}
\vspace{-4mm}
\end{figure}

We introduce FAILURE, a multimodal dataset, containing 229 real-world manipulation data recorded with a Baxter robot. The experimental setup is presented in Fig.~\ref{fig:baxter}. We use a Baxter humanoid robot with the following equipments: a parallel gripper, an Asus Xtion Pro RGB-D camera mounted on the head, a PSEye microphone mounted on the lower torso and the Baxter's default RGB camera mounted on the left arm which is also used to monitor manipulation execution. The object sets used in the experiments are shown in Fig.~\ref{fig:objects}.

During the data collection, the robot is tasked to execute a manipulation scenario, and all synchronized sensor readings (i.e., RGB/RGB-D image streams, audio waves) are then  simultaneously recorded.

Among the primitive manipulation types introduced in \cite{Worgotter2013ontology}, we mainly  work on the following $5$  types:  

\textit{push}, \textit{pick\&place}, \textit{pour}, \textit{place-in-container}, \textit{put-on-top}. The distribution of the  successful and failed trials for each manipulation   is presented in Table \ref{tbl:dataset}.

Sampled frames for each successful and failed case are visualized in Fig.~\ref{fig:dataset}. See the supplementary video showing the robot execution of sample manipulations.

Regarding data labeling, we have followed a goal oriented  approach. The following cases are labelled as failure since the manipulation goal may not be met: dropping the manipulated object, collision with another object, collapsing a structure, localizing incorrectly, spilling the poured content onto the table. Minor cases, such as slight changes in the intended target location and/or orientation, are still considered as success.

\begin{table}[!b]
\centering
\caption{Distribution of the  successful and failed executions in our multimodal robot manipulation dataset FAILURE. \label{tbl:dataset}}
\begin{tabular}{|r|c|c|c|}
\hline
\multicolumn{1}{|c|}{\textbf{Manipulation}} & \textbf{\#Successes}  & \textbf{\#Failures} & \multicolumn{1}{l|}{\textbf{Total}} \\ \hline
Push                    & 12  & 19   & 31   \\ \hline
Pick\&place             & 13  & 32   & 45   \\ \hline
Pour                    & 25  & 42   & 67   \\ \hline
Place-in-container      & 23  & 33   & 56   \\ \hline
Put-on-top              & 9   & 21   & 30   \\ \hline
\multicolumn{1}{|l|}{}  & 82  & 147  & 229  \\ \hline
\end{tabular}
\vspace{-4mm}
\end{table}

\subsection{Data Preprocessing}
\label{sec:dataprocessing}

\begin{figure}[!th]
\centering
    \includegraphics[width=0.9\linewidth]{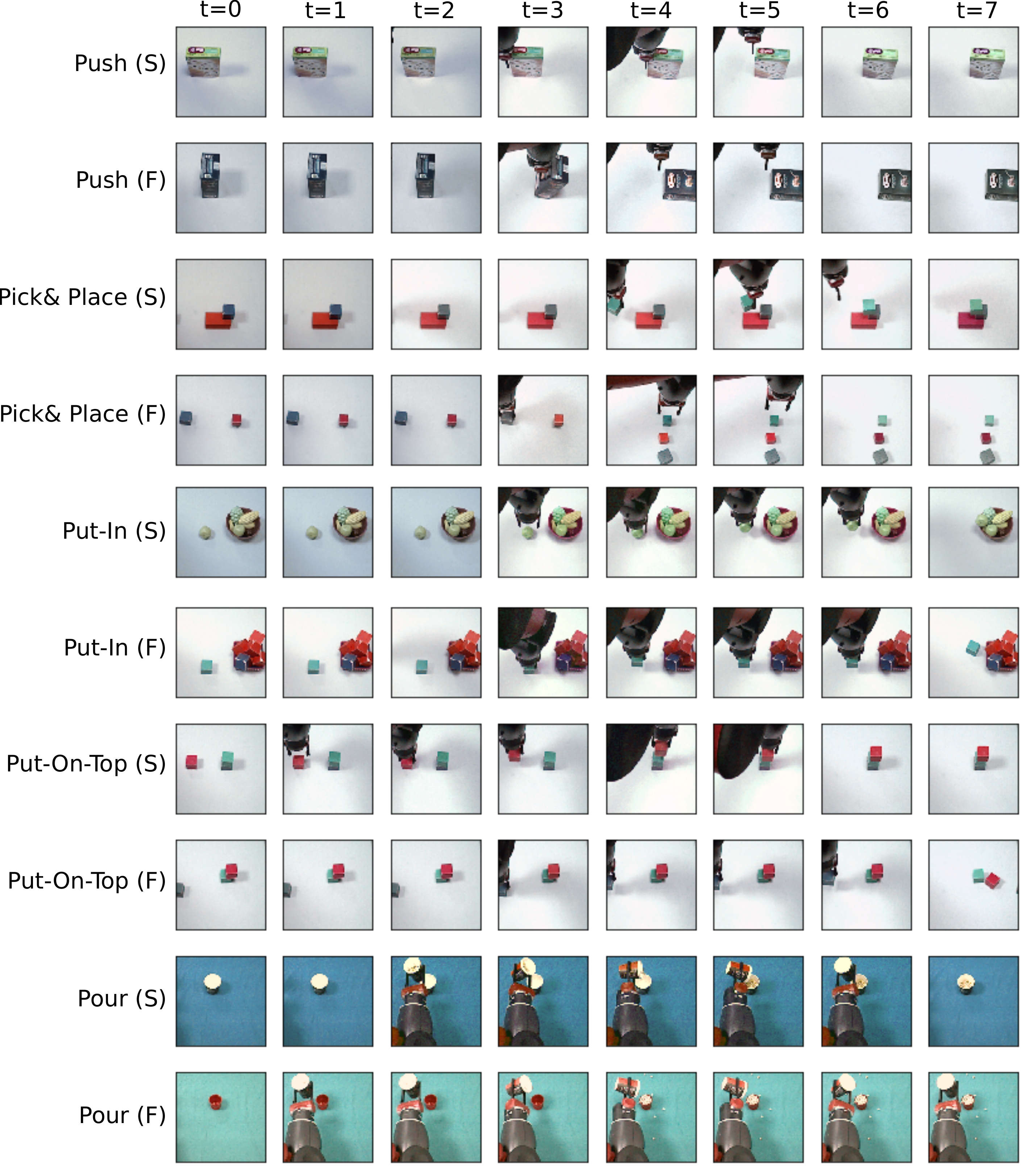}
    \caption{Sample successful and failed executions from the proposed FAILURE dataset. Higher resolution animated images can be found at \href{https://github.com/ardai/fino-net}{https://github.com/ardai/fino-net}.}
    \label{fig:dataset}
    \vspace{-4mm}
\end{figure}

The robot plans and executes trajectories online. The length of the 
recordings vastly vary due to differences in executions of the manipulation and trajectory types. During manipulation, the observability of the scene varies depending on the the robot's arm position. 
To automatically eliminate such occluded frames, a depth-based thresholding method is applied.
After filtering, remaining frames are 
roughly segmented into three phases corresponding to the \textit{approach}, \textit{manipulate}, and \textit{retreat} primitives, respectively. Next, we randomly sample  four frames from each of \textit{approach}  and \textit{retreat} primitives. 
Finally, the  $224\times224$ pixels area corresponding to the table plane is cropped. Fig.~\ref{fig:dataset} shows   randomly sampled $8$ frames for each manipulation type.

\begin{figure*}[!t]
\centering
    \includegraphics[width=0.9\linewidth]{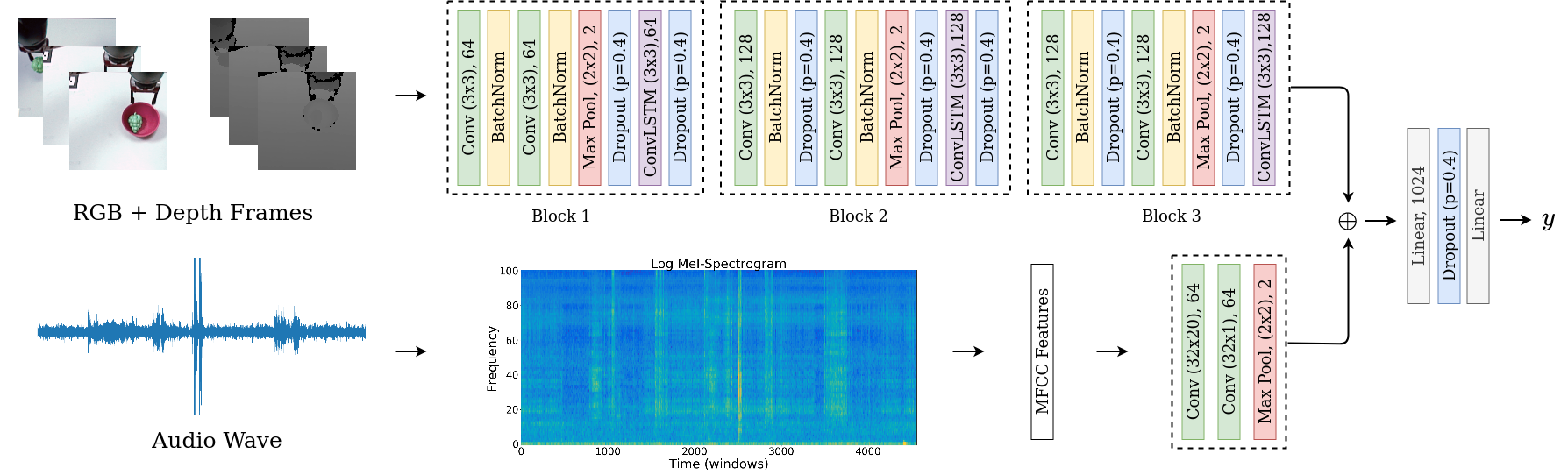}
    \caption{FINO-Net Architecture} 
    \label{fig:approach}
    \vspace{-5mm}
\end{figure*}

\subsection{FINO-Net}
\label{sec:approach}

To address the learning problem described in Section \ref{sec:problem_desc}, we propose FINO-Net as a multimodal classifier network to detect manipulation failures using onboard sensory data in real time. 
The inputs of the network are composed of RGB and depth frames captured from the head camera and audio waves recorded over the course of a robot manipulation. FINO-Net adopts early fusion to combine RGB and depth frames while applies  late fusion to combine vision and audio data. The architecture processes visual and audio inputs individually with a series of convolutional and convolutional-LSTM (convLSTM) layers. Finally, in the fusion step, the latent space representations are concatenated into a feature vector and fed to the fully connected layers. The overall FINO-Net architecture is depicted in Fig.~\ref{fig:approach}. In the following subsections, we elaborate more on the network architecture, loss function, and training details. 

\subsubsection{Vision}
In order to process spatio-temporal features in RGB and depth frames, we employ convLSTM cells. A typical LSTM cell is implemented using the fully connected layers, while a convLSTM replaces these with convolution operators.

The visual branch consists of three main blocks  (see the top branch in Fig.~\ref{fig:approach}). Inspired from \cite{Rothfuss2018deep}, each block is composed of two convolutional layers and a convLSTM layer.  
RGB and depth frames are early fused by stacking on top of each other before feeding to the first visual block.  
Inside each block, the filter numbers remain the same for all convolutional and convLSTM layers.  
Each convolutional layer has 3x3 filters. Before convLSTM layers, max pooling is applied to cut in half the number of features. Each block also has  batch normalization and dropout layers.

\subsubsection{Audition}
In our earlier works \cite{Inceoglu2018icra, Inceoglu2018iros} we present that audition can be employed to monitor manipulation execution as it provides complementary information to other sensing modalities. We also showed that Mel Frequency Cepstral Coefficients (MFCCs) are the suitable representations for auditory monitoring where Support Vector Machines were employed to classify audio features.

In this work, we adopt a convolutional network composed of two convolutional layers followed by a max pooling layer (see the bottom branch in Fig.~\ref{fig:approach}). There are 64 filters in each layer with a filter size of 32. As input, we use single channel audio recordings with 16 KHz sampling rate. The raw audio signal is divided into 32 millisecond windows. For each window, Short Time Fourier Transform is applied to convert signal into frequency domain. Mel filterbank is applied and 20 MFCCs are obtained using the librosa \cite{Mcfee2015librosa} library. The number of the audio windows are fixed by either applying padding or clipping.

\subsubsection{Fusion}
We adopt a late fusion approach to combine visual and auditory modalities. We introduce the following model:
\begin{equation}
y = \Phi(\phi_1(x_1^{(1)},...,x_{t_1}^{(1)}) \oplus ... \oplus \phi_m(x_1^{(m)},...,x_{t_m}^{(m)}))~,
\end{equation}

where $\phi_m$ is a unimodal convolutional neural network which acts as feature extractor, $\oplus$ is the concatenation operator, and $\Phi$ is the late fusion based classifier network. In the fusion step, vision and audition features, obtained from final output of each modality, are concatenated into a single feature vector. The fusion layer is composed of two fully connected layers as shown in Fig.~\ref{fig:approach}. There can be made connections between the proposed hierarchical fusion approach and primate visual cortex \cite{kruger2012deep}.

\subsubsection{Optimizer And Regularization} 
\label{sec:regularization}
As an optimizer, we use Adam with a learning rate of $1e-6$. 
Furthermore, batch normalization is applied after each convolutional layer. To boost the roles of very basic features (e.g., edges and curves), a central dropout approach is adopted with the probability rate of $0.4$. After each convolutional, convLSTM, and fully connected layers, a dropout layer is inserted except the first convolutional layer in Block 1  and the last convLSTM layer in Block 3.

To prevent overfitting, we augment the data by applying color augmentation and random flipping. For instance, the brightness, contrast, saturation, and hue values of all images in a sequence are randomly changed with a probability of $0.2$.
In a similar fashion, each image sequence is flipped vertically with a probability of $0.5$.

\section{Experiments}

We split our dataset  into train (70\%) and test (30\%) sets by preserving the class distribution. Experimental results are presented in terms of class weighted precision, recall and F1-scores. Reported results are the highest scores obtained after applying early stopping. 
\subsection{Baselines}
We compare FINO-Net with a VGG-based network~\cite{Simonyan2014}, which has 16 convolutional layers extended with an LSTM layer. The VGG baseline network is pretrained on ImageNet to be further used as a feature extractor. Obtained features from the final convolutional layer are fed into a single layer LSTM with 1024 neurons, followed by a fully connected layer. During training, only LSTM and fully connected layer weights are updated. For a fair comparison, this baseline model is trained with the same parameters (e.g. learning rate, batch size, etc.) and the same strategy (e.g., loss function, data augmentation, etc.) used for the training of FINO-Net. Note that the baseline model is also  extended with the same fusion structure that FINO-Net has. Following baseline networks are trained with the given modality data:

\begin{itemize}
    \item VGG-RGB: The input of the network is only the RGB frames obtained from the head camera.
    \item VGG-D: The input is only the single channel depth  (D) frames captured from the head camera.
    \item VGG-RGB-D: RGB and depth frames are stacked on top of each other to obtain a 4 channel RGB-D input.
    \item VGG-RGB-D-A: Similar to FINO-Net, stacked RGB-D frames and audio (A) features are first individually processed, and then are concatenated to be fed to the fully connected layer.
\end{itemize}

\subsection{FINO-Net}
To analyse  the  unique  contribution  of  each sensing modality on the failure detection 
task, we trained several variations of FINO-Net. 
To boost the performance, all convolutional and convLSTM layers of FINO-Net are initialized with the VGG weights pretrained on ImageNet. Next, we perform various training operations with the following modality data from our FAILURE dataset: 

\begin{itemize}
    \item FINO-Net-RGB: Network is trained with only the RGB input. 
    
    \item FINO-Net-D: Network is trained with the depth (D) frames only.
    
    \item FINO-Net-A: Only the audio (A) branch is trained. After the convolutional layers, there is a single fully-connected layer with 64 neurons.  
    
    \item FINO-Net-RGB-D: The visual branch of FINO-Net is trained by stacking the RGB and depth frames as input to the network.
    
    \item FINO-Net-RGB-D-A: The entire network  in Fig.~\ref{fig:approach} is trained with all the given modalities.
    The visual branch weights are initialized with FINO-Net-RGB-D and updated during training.
\end{itemize}

In addition to those experiments, we also freeze the FINO-Net weights initialized with that of the VGG baseline model. Next, our dataset is employed to train only the last convLSTM and the subsequent layers. Finally, the following experiments are performed:

\begin{itemize}
    \item FINO-Net-F-RGB: While training the visual branch with RGB input, the layers are frozen (F) except the final convLSTM and fully connected layers.
    
    \item FINO-Net-F-RGB-D: The vision branch is duplicated for the RGB and depth frames. The RGB branch weights are frozen, except for the final convLSTM layer. All layers dedicated to the depth frames are updated. Next, the RGB and depth features are concatenated and fed to the fully connected layers. 

    \item FINO-Net-F-RGB-D-A: In addition to the previous FINO-Net-F-RGB-D, the audio features are concatenated with visual features before passing to the fully connected layers. 
    
\end{itemize}

Note that in the experiments FINO-Net-F-RGB-D and  FINO-Net-F-RGB-D-A, we treat the RGB and depth channels separately by duplicating the visual branch. This is because VGG features are learned from the RGB ImageNet images, and thus, are not compatible with single-channel depth features. By updating the layers in the depth branch, we let the network explore unique depth cues which boost the network performance. This decoupling between the RGB and depth streams  can also be interpreted as  late fusion which is different than the early fusion strategy employed in FINO-Net-RGB-D and FINO-Net-RGB-D-A. 

Note also that since the VGG network  is particularly implemented for the structured image data, it cannot be trained with the audio modality. We, therefore, omit the model VGG-A and instead use FINO-Net-A for a fair comparison. The same applies to FINO-Net-F-A. Recalling the fact that VGG features learned from ImageNet images are not compatible with the depth features, we skip FINO-Net-F-D and instead use FINO-Net-D.

\newcommand{\ra}[1]{\renewcommand{\arraystretch}{#1}}
\begin{table}[!t]\centering
\caption{Quantitative Evaluation} 
\label{tab:quanresults}
\scalebox{0.85}{
\ra{1.3}
\begin{tabular}{@{}lccccccc@{}}\toprule 
& \multicolumn{3}{c}{Failure Detection} & \phantom{abc} \\
\cmidrule{2-4} \cmidrule{6-8}
& Precision & Recall & F1       \\ \midrule
FINO-Net-RGB      & 84.59  & 84.72  & 84.49   \\
FINO-Net-D        & 85.34  & 84.72  & 84.07   \\
FINO-Net-A        & 88.07  & 87.50  & 87.63   \\
FINO-Net-RGB-D    & 89.19  & 88.88  & 88.97    \\
FINO-Net-RGB-D-A  & 98.64  & 98.61  & \textbf{98.60} \\
\bottomrule
FINO-Net-F-RGB      & 86.13  & 86.11  & 85.82   \\
FINO-Net-F-RGB-D    & 87.01  & 86.11  & 86.29   \\
FINO-Net-F-RGB-D-A  & 95.90  & 95.83  & 95.84  \\
\bottomrule
VGG-RGB          & 90.39  & 90.27  & 90.31  \\
VGG-D            & 89.19  & 88.88  & 88.97  \\
VGG-RGB-D        & 90.39  & 93.27  & 90.31  \\
VGG-RGB-D-A      & 94.44  & 94.44  & 94.44  \\
\bottomrule
\end{tabular}
}
\end{table}

\subsection{Quantitative Results}
The obtained quantitative results for the test split are reported in Table \ref{tab:quanresults}. 
These results clearly show that 
our proposed FINO-Net has incremental performance improvement when a new sensor modality is introduced. Thanks  to the proposed combination of the early and late fusion approaches, the FINO-Net's accuracy on the RGB data significantly increases with the depth and audio features. The same applies to the FINO-Net-F model where the model weights are mostly frozen. 
Note that the contribution of the audio data (i.e., FINO-Net-RGB-D-A) is relatively much more than that of the depth cues (FINO-Net-RGB-D). It is because when a failure (such as an object fall down) occurs, there are observed abrupt changes in the sound recordings. Such changes are not explicitly obtained in the captured depth images which are randomly sampled as described in section~\ref{sec:dataprocessing}. 

When it comes to the baseline model VGG, we do not have the same observations: the depth modality (VGG-RGB-D) exhibits no contribution to the head camera data, i.e. VGG-RGB.
The contribution of the audio modality is still less compared to the one observed in both FINO-Net models (FINO-Net-RGB-D-A and  FINO-Net-F-RGB-D-A). These findings clearly show that our proposed FINO-Net has enough capacity to capture the unique contribution of each modality. 

In terms of accuracy, our proposed FINO-Net-RGB-D-A (as well as FINO-Net-F-RGB-D-A) considerably outperforms the baseline (VGG-RGB-D-A) by leading to the highest F1-score ($98.60\%$) in the failure detection scenarios where the task is to distinguish the successful and failed manipulation executions in a binary-classification manner.

\begin{table}[!b]
\caption{Ablation Study for the FINO-Net Design}
\label{tbl:ablation}
\begin{tabular}{lccc}
\multicolumn{1}{c}{\textbf{Approach}}          & \textbf{Precision} & \textbf{Recall} & \textbf{F1-Score} \\ \hline
FINO-Net-RGB-D-A  & 98.64  & 98.61  & 98.60 \\
-Without central dropout             & 94.44              & 94.44           & 94.44             \\
-Without batchnorm & 91,93              & 91.66           & 91.72             \\
-Without batchnorm and dropout & 88.07 & 87.50 & 87.63 \\
-Single layer audio branch        & 95.83              & 95.83           & 95.81             \\ \hline
\end{tabular}
\end{table}

Table \ref{tbl:ablation} summarizes the ablation study for the FINO-Net model design choices we have made. 
Comparing the results, the central dropout applied together with batch normalization layers leads to ~11\% performance boost. Furthermore, increasing the number of convolutional layers for the audio branch slightly increases the network performance.

\begin{figure}[!t]
\centering
    \includegraphics[width=0.6\linewidth]{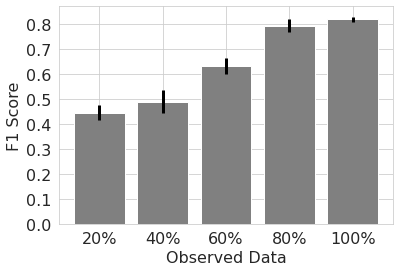}
    \caption{Fino-Net-RGB-D-A inference results on incomplete observations. For each recording in the test set, 10 different inferences are made with frames randomly sampled from the entire course of a manipulation.} 
    \label{fig:online_inference}
    \vspace{-3mm}
\end{figure}

Fig.~\ref{fig:online_inference} presents FINO-Net inference results in the case of having incomplete observations. Results indicate that observations right after the robot arm retreats from the scene, i.e., after $80\%$, are more informative. Therefore, as explained in section~\ref{sec:dataprocessing}, we prefer to sample frames from the \textit{retreat} phase, and to capture differences in the scene during the execution we also sample frames from the early \textit{approach} phase. In addition, Table \ref{tbl:runtime} presents the FINO-Net execution time for one single forward pass  obtained on Nvidia 2080 Ti GPU. The obtained results are average values of multiple executions. The results in Table \ref{tbl:runtime} and Fig.~\ref{fig:online_inference} show that even though the model runs in real-time, inferences are more accurate at the end of the execution.

\begin{table}[!b]
\centering
\caption{Inference Durations on Nvidia 2080 Ti GPU}
\label{tbl:runtime}
\resizebox{\linewidth}{!}{%
\begin{tabular}{lc|lc}
\hline
\textbf{Approach} & \multicolumn{1}{l|}{\textbf{Run time (msec)}} & \textbf{Approach} & \multicolumn{1}{l}{\textbf{Run time (msec)}} \\ \hline
FINO-Net-RGB & 9.4 & FINO-Net-A & 1 \\
FINO-Net-D & 9.3 & FINO-Net-RGB-D-A & 9.8 \\
FINO-Net-RGB-D & 9.3 & VGG-RGB-D-A & 3.4 \\ \hline
\end{tabular}%
}\vspace{-4mm}
\end{table}

\section{Discussion and Conclusion}

In this work, we presented a novel deep network model, named FINO-Net, that employs multi-model sensor readings to capture spatio-temporal features of robot manipulation executions to detect  possible failures. In addition, we introduced a new multimodal manipulation dataset FAILURE with failed robot execution attempts. We conducted various experiments and compared the  performance of FINO-Net with a VGG-based baseline model. 

Findings in Table~\ref{tab:quanresults} show that FINO-Net has enough capacity to capture unique contribution of each sensor modality, which is not the case for the baseline model. This plays a crucial role when the robot plans a recovery action to prevent such failures in time. For instance, the robot can autonomously decide what modality type(s) to rely on, in order to react to a failure with the most optimal action.

We are aware of the fact that our proposed dataset has limited number of samples. Therefore, we also plan to extend our dataset with more semantically different manipulations listed in \cite{Worgotter2013ontology}. Increasing the number of sensors by including  introspective sensors (i.e., force-torque and tactile) and robot hand cameras is another planned task in our agenda.

\bibliographystyle{IEEEtran}
\bibliography{references}  

\end{document}